\documentclass[sigconf]{acmart}

\AtBeginDocument{%
  }

\setcopyright{acmcopyright}

\copyrightyear{2022}
\acmYear{2022}
\setcopyright{acmcopyright}\acmConference[CIKM '22]{Proceedings of the 31st ACM International Conference on Information and Knowledge Management}{October 17--21, 2022}{Atlanta, GA, USA}
\acmBooktitle{Proceedings of the 31st ACM International Conference on Information and Knowledge Management (CIKM '22), October 17--21, 2022, Atlanta, GA, USA}
\acmPrice{15.00}
\acmDOI{10.1145/3511808.3557580}
\acmISBN{978-1-4503-9236-5/22/10}

\usepackage{amssymb,amsmath,amsfonts}
\usepackage{graphicx}
\usepackage{url}
\usepackage{color}
\usepackage{amsfonts}
\usepackage{blindtext}
\usepackage{mathrsfs}
\usepackage{tabularx}
\usepackage{bbm}
\usepackage{url}
\usepackage{paralist}
\usepackage[english]{babel}
\usepackage{booktabs}
\usepackage{multicol}
\usepackage{multirow}
\usepackage{subfigure}
\usepackage{amsmath}
\usepackage{graphicx}
\usepackage{CJKutf8}
\usepackage{enumitem}
\usepackage{balance}
\begin{document}
\begin{CJK}{UTF8}{gbsn}
\title{\textsc{DialogID}: A Dialogic Instruction Dataset for Improving Teaching Effectiveness in Online Environments}

%
% The "author" command and its associated commands are used to define
% the authors and their affiliations.
% Of note is the shared affiliation of the first two authors, and the
% "authornote" and "authornotemark" commands
% used to denote shared contribution to the research.
%\email{webmaster@marysville-ohio.com}
%\affiliation{%
%  \institution{Institute for Clarity in Documentation}
%  \streetaddress{P.O. Box 1212}
%  \country{USA}

\author{Jiahao Chen}
\affiliation{%
  \institution{TAL Education Group}
  \city{Beijing}
  \country{China}}
\email{chenjiahao@tal.com}

\author{Shuyan Huang}
\affiliation{%
  \institution{TAL Education Group}
  \city{Beijing}
  \country{China}}
\email{huangshuyan@tal.com}

\author{Zitao Liu}
\authornote{The corresponding author: Zitao Liu.}
\affiliation{%
  \institution{Guangdong Institute of Smart Education\\Jinan University}
  \city{Guangzhou}
  \country{China}}
\affiliation{%
  \institution{TAL Education Group}
  \city{Beijing}
  \country{China}}
\email{liuzitao@tal.com}

\author{Weiqi Luo}
\affiliation{%
  \institution{Guangdong Institute of Smart Education\\Jinan University}
  \city{Guangzhou}
  \country{China}}
\email{lwq@jnu.edu.cn}

%  \institution{The Th{\o}rv{\"a}ld Group}
%  \streetaddress{1 Th{\o}rv{\"a}ld Circle}

% \affiliation{%
%  \institution{Inria Paris-Rocquencourt}

% \affiliation{%
% \institution{Rajiv Gandhi University}

% \affiliation{%
%  \institution{Tsinghua University}
%  \streetaddress{30 Shuangqing Rd}
%\affiliation{%
%  \institution{Palmer Research Laboratories}
%  \streetaddress{8600 Datapoint Drive}
%  \country{USA}

\renewcommand{\shortauthors}{Jiahao Chen, Shuyan Huang, Zitao Liu, \& Weiqi Luo}


\begin{abstract}
\label{sec:abstract}
Online dialogic instructions are a set of pedagogical instructions used in real-world online educational contexts to motivate students, help understand learning materials, and build effective study habits. In spite of the popularity and advantages of online learning, the education technology and educational data mining communities still suffer from the lack of large-scale, high-quality, and well-annotated teaching instruction datasets to study computational approaches to automatically detect online dialogic instructions and further improve the online teaching effectiveness. Therefore, in this paper, we present a dataset of online dialogic instruction detection, \textsc{DialogID}, which contains 30,431 effective dialogic instructions. These teaching instructions are well annotated into 8 categories. Furthermore, we utilize the prevalent pre-trained language models (PLMs) and propose a simple yet effective adversarial training learning paradigm to improve the quality and generalization of dialogic instruction detection. Extensive experiments demonstrate that our approach outperforms a wide range of baseline methods. The data and our code are available for research purposes from: \url{https://github.com/ai4ed/DialogID}.
\end{abstract}




\maketitle

\section{Introduction}
\label{sec:intro}
The Covid-19 pandemic has brought tremendous changes to educational institutions around the world. With the recent development of technology such as digital video processing and live streaming, various forms of online learning tools emerge and a large number of offline institutions switch to the online mode \cite{dhawan2020online,li2020multimodal,liu2020dolphin}. In spite of the advantages of online classes and a variety of support from online teaching software, teaching online classes still remains a very challenging task for the well-trained offline classroom instructors. When sitting in front of a camera or a laptop, traditional classroom instructors lack effective pedagogical instructions to ensure the overall quality of their online classes. 

Dialogic instructions for online classes promote interactions between teachers and students instead of teacher-presentation only. Besides, they improve the learning interest and the confidence of the student, constructing effective learning habits. Hence, a computational approach to automatically detect the dialogic instructions during the online class seems to provide real-time feedback to teachers and improve their online teaching skills. 

However, there are some challenges to build an automatic detection approach of dialogic instruction. Online teaching is not a standardized procedure. Even for the same learning content, the instructors teach varies according to their own pedagogical styles. Furthermore,  the different teaching experiences of the instructors also lead to the quality of dialogic instructions. An illustrative example of \emph{note-taking}\footnote{Note-taking instructions ask students to take notes of key points.} instructions is as follows:

\begin{itemize}
	\item S1: Make sure you \emph{write down} this key point.
	\item S2: Make sure you \emph{remember} this key point. 
\end{itemize}

S1 instruction gives students concrete action, i.e., taking notes, which is helpful for students to build their learning habits. S2 is an ineffective and confusing instruction and doesn't meet the quality standard of note-taking instruction. An intelligent dialogic instruction detection model is hoped to effectively distinguish these subtle differences and provide instant feedback to online instructors. %Developing such AI driven approaches requires a large-scale, high-quality, and well-annotated teaching instruction datasets to cover various teaching styles and pedagogical scenarios.

\begin{table*}[!tbph]
	\centering
	\small
	\setlength \tabcolsep{3pt}
	\caption{\label{tab:instructions}Definitions and examples of dialogic instructions. \textit{others} contains instructions that are either ineffective or irrelevant.} \vspace{-0.2cm}
	\begin{tabularx}{17cm}{lXX}
		\toprule
		\textbf{Instruction} & \textbf{Definition}                                                                                                                                           & \textbf{Example(s)}         \\
		\midrule
		commending           & Commending instructions that praise and encourage students.                                              & Good job!                              \\
		guidance             & Guiding students to solve a problem step by step.                                    & What would happen then?                             \\
		summarization        & Wrapping up the lesson or summarizing the content just learned.                                                       & Let’s conclude what we have learned today.          \\
		greeting             & Greetings at the beginning of a class; Instructions that help manage the teaching procedures. & How is it going? \newline Can you see the slides?               \\
		note-taking          & Instructions that ask students to take notes of key points.                                                           & Make sure you write down this key point.          \\
		repeating            & Requiring students to rehearse the content.                      & Could you repeat it?                 \\
		reviewing            & Reminding the students what they learned in a previous class.                                                          & Could you remember the words you learned last week? \\
		example-giving       & Demonstrating the content by concrete facts.                                                                             & Here is an example.                                 \\
		others          & Ineffective instructions, or instructions unrelated to the class.                                     & It's good weather today.          \\
		\bottomrule    
	\end{tabularx}
	\vspace{-0.2cm}
\end{table*}

Existing educational research has revealed the significance of dialogic instructions on students' social emotional well-being \cite{tennant2015students}, motivation \cite{henderlong2002effects}, and academic achievements \cite{moely1992teacher,dweck2007boosting}. Class observation frameworks have been established such as \textit{CLASS} \cite{pianta2008classroom} and \textit{COPUS} \cite{smith2013classroom}. However, these methods heavily rely on human efforts like manual video coding \cite{praetorius2018classroom,rosenshine2012principles}, and hence fail to provide automatic in-time feedback to instructors. Machine learning models are able to learn from human-coded data then make predictions automatically. For example, Donnelly et al. utilized Naive Bayes models to capture the occurrences of five key instructional segments, i.e., small group work, lecture, etc. \cite{donnelly2016automatic}. 

However, even though the aforementioned research focuses on detecting and studying teachers' dialogic instructions, none of them open sources their research datasets. Furthermore, the majority of research works are undertaken in the traditional offline classrooms and their methodologies and paradigms are not applicable to the online learning environments. Therefore, in this work, to help and promote research and development of tasks for online dialogic instruction detection, we present \textsc{DialogID}, a high-quality dialogic instruction dataset for improving online teaching effectiveness. \textsc{DialogID} contains 30,431 effective dialogic instructions extracted from real-world K-12 online classes. To the best of our knowledge, \textsc{DialogID} is one of the first publicly available dialogic instruction datasets collected from online classrooms. Furthermore, we propose a simple yet effective adversarial training (AT) paradigm with pre-trained language models (PLMs) learned from \textsc{DialogID} to solve the dialogic instructions detection problem automatically. Experimental results demonstrate the usage and effectiveness of the \textsc{DialogID} dataset and the proposed instructions detection approach.

% \vspace{-0.1cm}
\section{Dataset}
\label{sec:dataset}
\subsection{Dialogic Instructions}
In this work, following many existing pedagogical studies \cite{goodenow1993psychological,osterman2010teacher,henderlong2002effects,dweck2007boosting,yelland2007rethinking,shafto2014rational,anthony2015supporting,an2004capturing,haghverdi2010note,lee2008effects,rinehart1986some}, we focus on online dialogic instructions with the following aspects: (1) motivate students and make them feel easy about the class: \textit{greeting} \cite{goodenow1993psychological,osterman2010teacher} and \textit{commending} \cite{henderlong2002effects,dweck2007boosting}; (2) help students understand learning materials and retain them: \textit{guidance} \cite{yelland2007rethinking}, \textit{example-giving} \cite{shafto2014rational}, \textit{repeating} \cite{anthony2015supporting}, and \textit{reviewing} \cite{an2004capturing}; and (3) build effective learning habits:  \textit{note-taking} \cite{haghverdi2010note,lee2008effects} and \textit{summarization} \cite{rinehart1986some}. 

Therefore we aim to capture these 8 kinds of effective instructions. The definitions and examples of instructions are shown in Table \ref{tab:instructions}. Please note that the scope of dialogic instructions in our work is a superset of the previous study \cite{xu2020automatic}.

\subsection{Data Annotation}
To ensure the annotation quality and the trained AI driven detection models are able to be deployed into the real production systems without any human intervention, we design a 3-step online dialogic instruction annotation process that aims to automatically identify teaching instructions from the entire online classroom recordings. The 3-step process is described as follows.

\noindent \textbf{Step 1: Extract teacher utterances}. Similar to \cite{xu2020automatic,huang2020neural}, we extract teacher utterances from the online classroom video recordings and filter out background noises and silence fragments via an in-house voice activity detection (VAD) model. Similar to \cite{tashev2016dnn}, the in-house VAD model is a four-layer deep neural network trained on online classroom audio data. Please note that there are no voice overlaps as audio recordings of each teacher and student are recorded separately.

\noindent \textbf{Step 2: Generate dialogic instruction candidates}. Dialogic instructions only constitute a small portion of teacher utterances within an online course. To make the annotation efficiently and economically, we identify utterance candidates that may contain dialogic instructions. Specifically, we first transcribe each teacher utterance (obtained from Step 1) via a self-trained automatic speech recognition (ASR) model, which is a deep feed-forward sequential memory network to transfer the voice utterances into text information \cite{zhang2018deep}. The ASR model is trained on classroom specific datasets and has a character error rate of 11.36\% in the classroom scenarios. Then for each type of dialogic instruction listed in Table \ref{tab:instructions}, we pre-define a list of keywords and use the keyword matching method to find candidate utterances of dialogic instructions. Only the utterance whose transcription is matched with at least one of the keywords will be kept. The pre-defined keywords are constructed by analyzing thousands of online class videos and surveying hundreds of instructors, students, parents and educators. For example, words or phrases like ``Hello/Good Morning/Goodbye'' and ``as seen in/as shown in'' are  keywords for the greeting and summarization dialogic instructions respectively.

\noindent \textbf{Step 3: Extract segment-level audios for utterance-level annotation}. Individual utterance candidates from Step 2 may simply contain one or two sentences that are difficult to annotate due to the lack of classroom contexts. Therefore, to make sure our crowdsourced labels are reliable, we assemble the target utterance candidate, its preceding \emph{n} utterances and its \emph{n} following utterances into an audio segment. The crowd workers assign labels after listening to each audio segment. A teacher's utterance will be labeled as ``others'' if it doesn't belong to the 8 categories in Table \ref{tab:instructions}.

\subsection{Data Analysis}
\begin{figure}[!htbp]
\footnotesize
\centering
\includegraphics[width=0.47\textwidth]{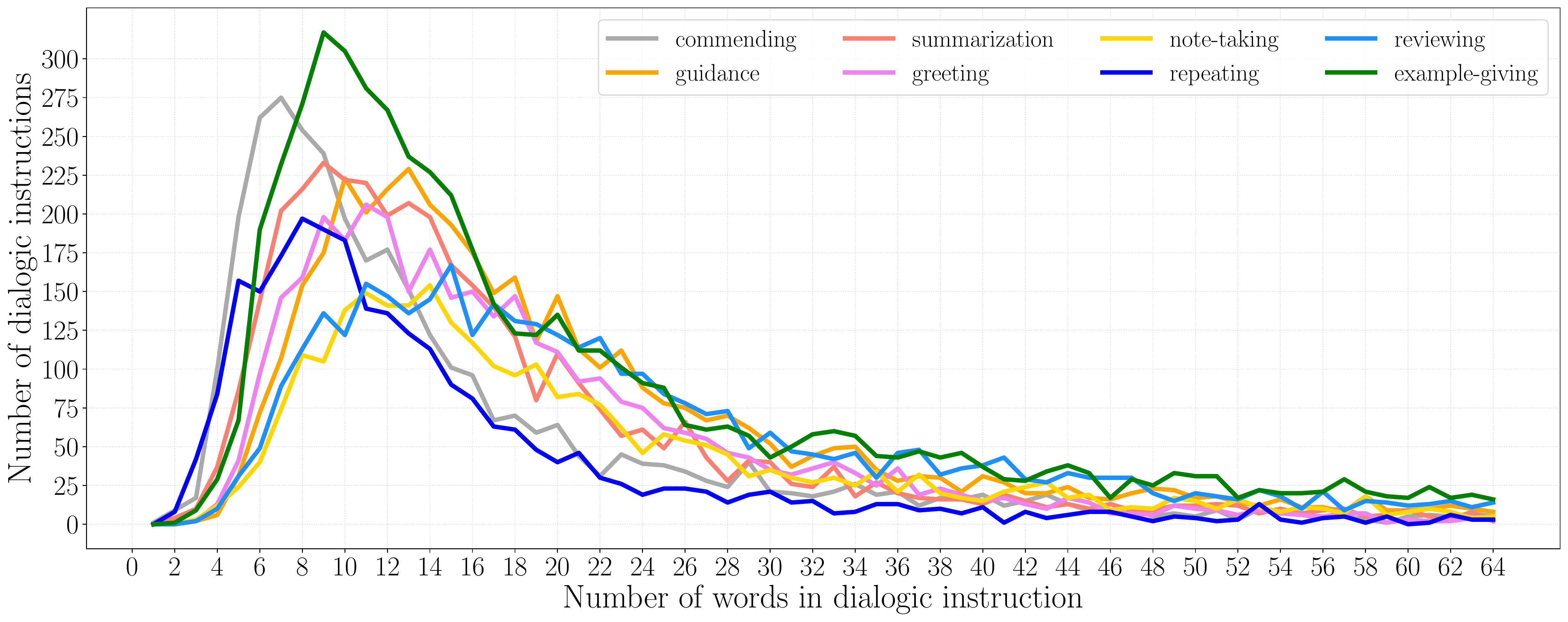} \vspace{-0.3cm}
\caption{Length distribution per type in \textsc{DialogID}.}
\Description{Length distribution per type}
\label{fig:len_distribution}
\vspace{-0.35cm}
\end{figure}

Dialogic instructions in DialogID are collected and constructed from the K-12 online classes at TAL Education Group\footnote{TAL Education Group (NYSE:TAL) is an educational technology company dedicated to supporting public and private education across the world.}. Through the 3-step annotation procedure, we end up with 51,908 annotated samples and 30,431 of them are effective online dialogic instructions. The detailed per type instruction distribution and the corresponding sizes of training, validation, test sets are shown in Table \ref{tab:statistics}. Furthermore, the length distribution (in words) of each type of dialogic instruction is depicted in Figure \ref{fig:len_distribution}. As we can see, the amounts of different types of dialogic instructions are relatively balanced in \textsc{DialogID} and most of the dialogic instructions are short sentences with less than 20 words.

\begin{table}[!htbp]
\centering
\footnotesize
\caption{Data statistics of the \textsc{DialogID} dataset.} \vspace{-0.3cm}
\label{tab:statistics}
\setlength{\tabcolsep}{1.2mm}{
\begin{tabular}{lcccrcc}
\toprule
               & \textbf{Train} & \textbf{Validation} & \textbf{Test} & \textbf{Total}& \textbf{AVG Len}& \textbf{STD Len}  \\
\midrule
commending     & 2,437     & 320        & 692     & 3,449                      & 17.2       & 16.3       \\
guidance       & 2,987     & 425        & 881     & 4,293                      & 23.4       & 18.4       \\
summarization  & 2,206     & 307        & 588     & 3,101                      & 27.1       & 23         \\
greeting       & 1,798     & 243        & 529     & 2,570                      & 15.5       & 13.8       \\
note-taking    & 2,667     & 394        & 782     & 3,843                      & 19         & 15.3       \\
repeating      & 2,488     & 368        & 705     & 3,561                      & 19.9       & 14.2       \\
reviewing      & 2,793     & 402        & 786     & 3,981                      & 27.3       & 19.6       \\
example-giving & 3,977     & 550        & 1,106    & 5,633                      & 24.8       & 20.5       \\
others         & 14,982    & 2,182       & 4,313    & 21,477                     & 21.6       & 17.1       \\
Total          & 36,335    & 5,191       & 10,382   & 51,908                     & 22         & 18         \\
\bottomrule
\end{tabular}}
\vspace{-0.4cm}
\end{table}

\section{An Adversarial Training Enhanced Detection Framework}
\label{sec:method}
In this section, we describe our dialogic instruction detection framework, which has two key components: (1) a PLM, which serves as the base model in the classification task; and (2) an adversarial training learning module, which improves the model generalization from the very limited and noisy teacher instruction transcriptions.

\subsection{Pre-trained Language Models}

Traditional machine learning models use word vectors as inputs, which are static and not able to extract contextual information. By contrast, more recent PLMs learn contextual embeddings with their Transformer-based architectures. Therefore in this study we utilize PLMs as our base model in our detection framework.

To perform the instruction detection task on a sentence $ \mathbf{x} = (x_1, \cdots, x_n) $ which contains $n$ tokens, similar to \cite{devlin2019bert,liu2019roberta}, we first add a special token $[CLS]$ in front of the sentence. After that, embeddings of each token $(E_{[CLS]}, E_1, \cdots, E_n)$ are fed into multiple Transformer encoders sequentially, where each token gradually captures contextual information of the sentence. Finally at the last layer of Transformer encoders, hidden states of each token are extracted, and the hidden state of the special token $[CLS]$  is treated as the representation of the sentence.

In our study, we utilize RoBERTa \cite{liu2019roberta} pre-trained model, which is a Transformer-based model sharing the same structure with BERT \cite{devlin2019bert}, while several improvements are made at the pre-training stage, including removing the next-sentence prediction objective of BERT, and using dynamic masking, etc. We also experiment with other recently proposed PLMs and details are discussed in Section \ref{sec:experiments}.

\subsection{Adversarial Training Module}

Adversarial training is an efficient regularization technique that provides an efficient way to not only improve the robustness of the DNNs against perturbations but also enhance its generalization over original inputs by training DNNs to correctly classify both of the original inputs and adversarial examples (AEs) \cite{miyato2017adversarial,goodfellow2014explaining,guo2021enhancing}. Similar to the pioneer work of Miyato et al. who extended AT to text classification \cite{miyato2017adversarial}, we create AEs by adding adversarial perturbations on intermediate representations in the embedding layers and use AEs to optimize the model parameters for better generalization. Specifically, the adversarial perturbation $\mathbf{e}$ is computed by an efficient and fast gradient approximation method developed by Goodfellow et al. \cite{goodfellow2014explaining} as follows:

\vspace{-0.3cm}
$$\mathbf{x}' = \mathbf{x} + \mathbf{e}; \quad \mathbf{e} = \epsilon \mathbf{g} /\|\mathbf{g}\|_{2}; \quad \mathbf{g} =\nabla_{\mathbf{x}} \mathcal{L}(\mathbf{x}, \boldsymbol{\theta})$$

 \noindent where $\mathbf{x}'$ and $\mathbf{x}$ denote the perturbed and original representations in neural networks' embedding layers respectively. $\epsilon$ is a hyperparameter to control the norm of perturbations and $\boldsymbol{\theta}$ is the model parameters.

\begin{table*}[!hbpt]
\centering
\caption{Prediction performance per instruction type of all different baselines in terms of precision, recall and F1 score.} \vspace{-0.3cm}
\footnotesize
\label{tab:performance}
\begin{tabular}{l|lccc|l|lccc}
\toprule
Dialogic Instruction                    & Model       & Precision       & Recall          & F1              & Dialogic Instruction                     & Model       & Precision       & Recall          & F1              \\
\hline
\multirow{6}{*}{commending}    & BERT        & 0.8274          & 0.8801          & 0.8529          & \multirow{6}{*}{guidance}       & BERT        & 0.7505          & 0.8025          & 0.7756          \\
                               & ELECTRA     & 0.8093          & 0.8829          & 0.8445          &                                 & ELECTRA     & 0.7518          & 0.8082          & 0.7790          \\
                               & MacBERT     & 0.8219          & 0.8801          & 0.8500          &                                 & MacBERT     & \textbf{0.8106} & 0.7480          & 0.7780          \\
                               & XLNet       & \textbf{0.8343} & 0.8223          & 0.8282          &                                 & XLNet       & 0.7899          & 0.7809          & 0.7854          \\
                               & RoBERTa     & 0.8013          & \textbf{0.9263} & 0.8592          &                                 & RoBERTa     & 0.7555          & 0.8241          & 0.7883          \\
                               & RoBERTa+AT & 0.8083          & \textbf{0.9263} & \textbf{0.8633} &                                 & RoBERTa+AT & 0.7770          & \textbf{0.8343} & \textbf{0.8046} \\
\hline
\multirow{6}{*}{summarization} & BERT        & \textbf{0.9039} & 0.8963          & 0.9001          & \multirow{6}{*}{greeting}       & BERT        & 0.8942          & 0.8790          & 0.8866          \\
                               & ELECTRA     & 0.8542          & 0.9167          & 0.8843          &                                 & ELECTRA     & 0.8392          & 0.8979          & 0.8676          \\
                               & MacBERT     & 0.8882          & \textbf{0.9184} & \textbf{0.9030} &                                 & MacBERT     & 0.8826          & 0.8809          & 0.8817          \\
                               & XLNet       & 0.8938          & 0.8878          & 0.8908          &                                 & XLNet       & 0.8248          & 0.9168          & 0.8684          \\
                               & RoBERTa     & 0.8834          & 0.9150          & 0.8989          &                                 & RoBERTa     & \textbf{0.9018} & 0.8507          & 0.8755          \\
                               & RoBERTa+AT & 0.8834          & 0.9150          & 0.8989          &                                 & RoBERTa+AT & 0.8637          & \textbf{0.9225} & \textbf{0.8921} \\
\hline
\multirow{6}{*}{note-taking}   & BERT        & 0.8100          & 0.9488          & 0.8740          & \multirow{6}{*}{repeating}      & BERT        & \textbf{0.9134} & 0.9277          & \textbf{0.9205} \\
                               & ELECTRA     & 0.8082          & 0.9373          & 0.8680          &                                 & ELECTRA     & 0.8728          & 0.9348          & 0.9027          \\
                               & MacBERT     & 0.7940          & \textbf{0.9514} & 0.8656          &                                 & MacBERT     & 0.8908          & \textbf{0.9376} & 0.9136          \\
                               & XLNet       & 0.8491          & 0.8632          & 0.8561          &                                 & XLNet       & 0.8770          & 0.9305          & 0.9030          \\
                               & RoBERTa     & 0.8201          & 0.9501          & \textbf{0.8803} &                                 & RoBERTa     & 0.8787          & 0.9248          & 0.9012          \\
                               & RoBERTa+AT & \textbf{0.8493} & 0.8939          & 0.8710          &                                 & RoBERTa+AT & 0.9006          & 0.9248          & 0.9125          \\
\hline
\multirow{6}{*}{reviewing}     & BERT        & 0.8162          & \textbf{0.9720} & 0.8873          & \multirow{6}{*}{example-giving} & BERT        & 0.9114          & 0.9675          & 0.9386          \\
                               & ELECTRA     & 0.8284          & 0.9644          & 0.8912          &                                 & ELECTRA     & 0.9066          & 0.9738          & 0.9390          \\
                               & MacBERT     & 0.8284          & 0.9644          & 0.8912          &                                 & MacBERT     & 0.9109          & 0.9702          & 0.9396          \\
                               & XLNet       & 0.8315          & 0.9542          & 0.8886          &                                 & XLNet       & 0.9033          & \textbf{0.9801} & 0.9402          \\
                               & RoBERTa     & 0.8346          & 0.9631          & 0.8943          &                                 & RoBERTa     & 0.9108          & 0.9792          & \textbf{0.9438} \\
                               & RoBERTa+AT & \textbf{0.8412} & 0.9567          & \textbf{0.8952} &                                 & RoBERTa+AT & \textbf{0.9126} & 0.9729          & 0.9418          \\
\bottomrule
\end{tabular}
\vspace{-0.3cm}
\end{table*}

\section{Experiments}
\label{sec:experiments}
To comprehensively assess \textsc{DialogID} and the proposed method, except for RoBERTa, we additionally select a series of widely-used text classification models, including BERT \cite{devlin2019bert}, ELECTRA \cite{clark2020electra}, MacBERT \cite{cui2020revisiting}, and XLNet \cite{yang2019xlnet}. Moreover, we conduct an ablation study to demonstrate the performance improvement enhanced by the adversarial training module. The proposed AT enhanced approach is denoted as ``RoBERTa+AT'' in the following section. Please note that the AT module can be incorporated with any PLM. PLMs for our experiments can be found in this repository \footnote{https://github.com/ymcui/Chinese-BERT-wwm}. For each model, we set max\_len to 128 and learning rate to 1e-5. The number of epochs is set to 100, and we early stop in the training phase if the model doesn't get improvement in validation datasets for 5 epochs.

\subsection{Results}

\noindent \textbf{Prediction with PLMs}. We compare the performance of different PLMs in terms of precision, recall and F1 scores. Results are shown in Table \ref{tab:performance} (per type) and Table \ref{tab:performance_overall} (overall). When comparing RoBERTa to other PLMs such as BERT, ELECTRA, MacBERT, and XLNet, we can find that RoBERTa achieves the highest prediction performance than other approaches in terms of F1 score, which indicates their stronger capacity to model dialogic instructions. When looking into each category, interestingly we can observe that RoBERTa does not always achieve the top performance. In detail, they show inferior performance compared with BERT and MacBERT in \textit{summarization}, \textit{greeting} and \textit{repeating}, which have the least samples compared to instructions in other categories.

\begin{table}[!hbpt]
\centering
\vspace{-0.2cm}
\caption{Overall prediction performance of different models.} \vspace{-0.2cm}
\footnotesize
\label{tab:performance_overall}
\begin{tabular}{l|ccc}
\toprule
Model       & Precision       & Recall          & F1  \\
\hline
BERT        & 0.8534          & 0.9092          & 0.8795         \\
ELECTRA     & 0.8338          & 0.9145          & 0.8720          \\
MacBERT     & 0.8534          & 0.9064          & 0.8779          \\
XLNet       & 0.8505          & 0.8920          & 0.8701          \\
RoBERTa     & 0.8483          & 0.9167          & 0.8802          \\
RoBERTa+AT & \textbf{0.8545} & \textbf{0.9183} & \textbf{0.8849}  \\
\bottomrule
\end{tabular}
\vspace{-0.2cm}
\end{table}

\noindent \textbf{Prediction with PLMs and AT}. We demonstrate the effectiveness of AT by comparing RoBERTa+AT with RoBERTa. Table \ref{tab:performance} and Table \ref{tab:performance_overall} show that by adding an adversarial training module which aims to enhance model's generalization, the RoBERTa+AT model outperforms the original RoBERTa model in 5 out of 8 types of dialogic instructions as well as the overall performance, in terms of F1 scores. It's worth noting that the RoBERTa+AT model increases F1 score by 1.66\% in \textit{greeting} compared with RoBERTa. We believe this is because the greeting dialogic instruction is the least category and it has the smallest average length compared to instructions in other categories. The AT module enhances the generalization of the PLM model by jointly training the original clean inputs and the corresponding perturbed AEs.

\begin{figure}[!thpb]
\vspace{-0.4cm}
\includegraphics[width=0.8\linewidth]{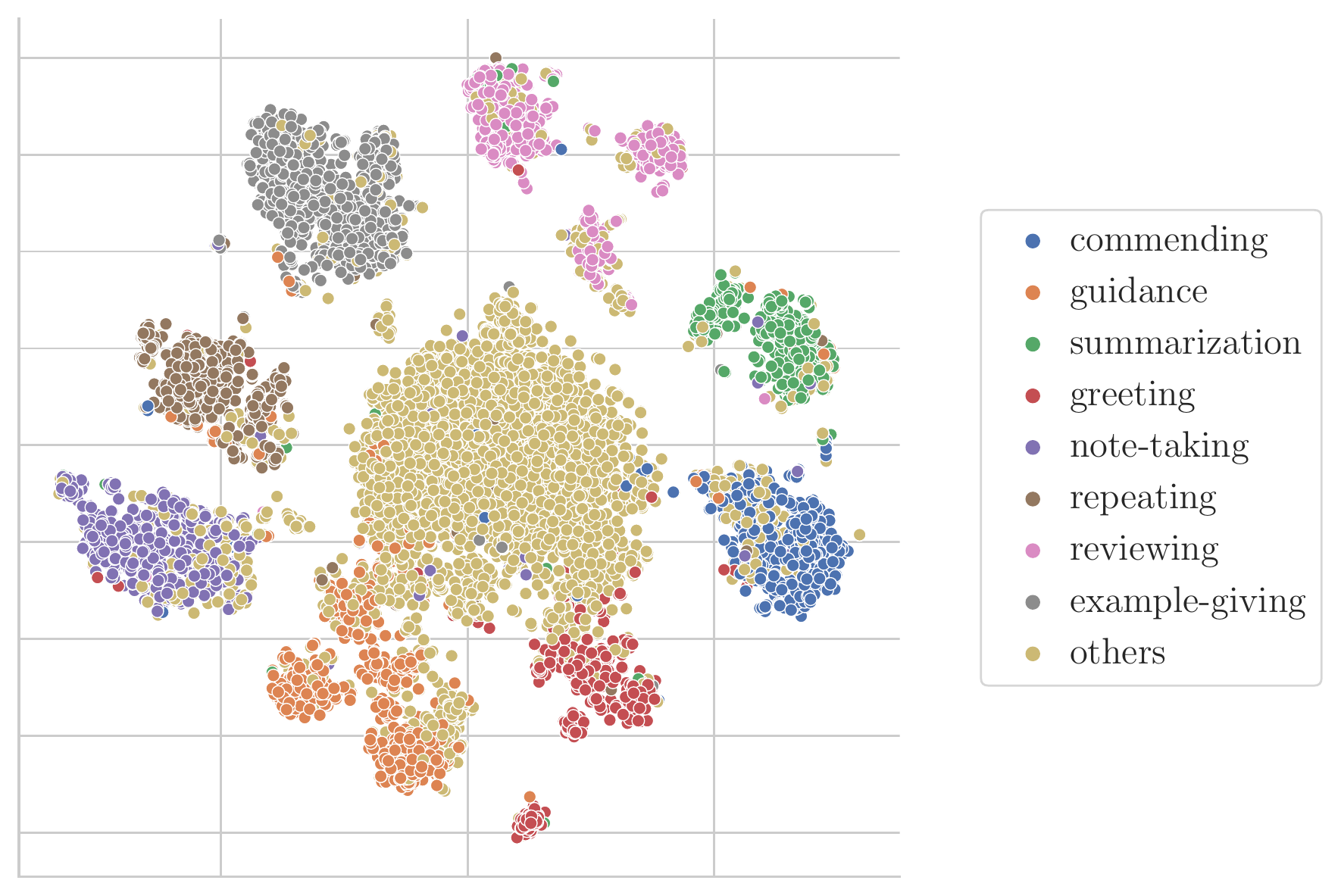} \vspace{-0.2cm}
\caption{Representation visualization of RoBERTa+AT.} 
\Description{Learned representation of the RoBERTa+AT model.} \vspace{-0.4cm}
\label{fig:visualize}
\end{figure}

\noindent \textbf{Qualitative Analysis}. We further demonstrate the effectiveness of RoBERTa+AT by visualizing the learned representations, shown in Figure \ref{fig:visualize}. Instances of nine categories in the testing set are fed into the trained models, and their representations i.e., the tensor values of the special token $[CLS]$ are collected. Dimension reduction method t-SNE \cite{van2008visualizing} is performed so that the representations can be visualized in the 2-dimensional space. From the figure, we can see that representations of instances in different categories are well separated by the proposed RoBERTa-AT with significant margins between different categories.

\section{Conclusion}
\label{sec:conclusion}

In this work, we introduce \textsc{DialogID}, a dialogic instruction dataset that contains 8 categories of the online class instructions collected from real-world K-12 online classrooms. Experiments conducted on \textsc{DialogID} show the effectiveness and superiority of our proposed approach against a wide range of baselines.

\begin{acks}
This work was supported in part by National Key R\&D Program of China, under Grant No. 2020AAA0104500; in part by Beijing Nova Program (Z201100006820068) from Beijing Municipal Science \& Technology Commission and in part by NFSC under Grant No. 61877029.
\end{acks}

\bibliographystyle{ACM-Reference-Format}
\balance
\bibliography{cikm2022}
\end{CJK}
\end{document}